\documentclass{article}

 \usepackage[preprint]{neurips_2026}

\usepackage[utf8]{inputenc} 
\usepackage[T1]{fontenc}    
\usepackage{hyperref}       
\usepackage{url}            
\usepackage{booktabs}       
\usepackage{amsfonts}       
\usepackage{nicefrac}       
\usepackage{microtype}      
\usepackage{xcolor}         
\usepackage{natbib}
\usepackage{graphicx}
\usepackage{subcaption}
\usepackage{wrapfig}
\usepackage{multicol}
\usepackage{multirow}
\usepackage{enumitem}
\usepackage{dblfloatfix}
\usepackage{bbm}
\usepackage{authblk}

\usepackage{amsmath}
\usepackage{amssymb}
\usepackage{mathtools}
\usepackage{amsthm}

\usepackage[capitalize,noabbrev]{cleveref}

\setcitestyle{square,numbers}
\title{Adapting Foundation Vision-Language Models to Medical Diagnosis via Query-Driven Expert Bridging}

\author[1,2]{Yitong Li$^*$}
\author[1,2]{Morteza Ghahremani$^*$}
\author[1,2]{Christian Wachinger}
\affil[1]{Lab for AI in Medical Imaging, Technical University of Munich (TUM), Germany}
\affil[2]{Munich Center for Machine Learning (MCML), Germany}

\begin{document}

\maketitle

\def\thefootnote{*}\footnotetext{Equal Contribution.}

\vspace{-1.0em}

\begin{abstract}
Vision-language foundation models achieve promising performance in natural image classification, yet their direct application to medical imaging is limited by severe domain shifts, resolution mismatches, and the multi-label nature of clinical diagnosis. 
Training dedicated medical foundation models from scratch, however, is costly and data-intensive. 
Here, we propose MedBridge, a lightweight adaptation framework that opens a new direction in domain-gap mitigation by jointly combining domain alignment, resolution preservation, and multi-label reasoning via complementary VLM experts for medical image diagnosis.
Specifically, MedBridge transforms pretrained VLMs into multi-view query encoders that inject a compact set of learnable query tokens into intermediate layers, enabling non-destructive domain alignment while preserving fine-grained pathological cues via multi-view high-resolution sampling. These query tokens further act as routing signals for a mixture-of-experts, dynamically integrating heterogeneous foundation models for multi-label reasoning without requiring a shared representation space.
We evaluated MedBridge on five chest radiograph benchmarks in three key adaptation tasks. MedBridge demonstrates superior performance in both cross-domain generalization (out-of-distribution transfer) and in-domain specialization (same-distribution tuning) settings, yielding a significant 6–15\% AUC improvement over state-of-the-art adaptation methods for multi-label thoracic disease diagnosis. Furthermore, MedBridge is model-agnostic and demonstrates broad extensibility across eight diverse VLMs (e.g., CLIP, LLaVA, Qwen-VL, MedGemma), highlighting its ability to flexibly adapt arbitrary foundation models into a powerful medical diagnostic tool. 
Our code will be released upon acceptance.
\end{abstract}

\newcommand{\conf}[1]{\;{\tiny(#1)}}
 

\vspace{-1.1em}
\section{Introduction}
\label{sec:intro}
\vspace{-0.7em}

Foundation vision-language models (VLMs) have gained substantial capabilities through large-scale pre-training on extensive datasets covering both visual and textual modalities~\cite{radford2021clip,cherti2023openclip,zhai2023siglip,liu2023llava,liu2024deepseek,yang2024qwen2}. This pre-training paradigm enables them to learn rich, transferrable cross-modal representations essential for a wide variety of downstream tasks~\cite{awadalla2023openflamingo,Yamaguchi_CVPR25_CLIP-Refine,tan2025vision,Jia2021align}. 
As these VLMs continue to grow in size, full fine-tuning of all parameters for specific downstream tasks becomes computationally intensive and elevates the risk of overfitting. Thus, recent research has focused on developing parameter-efficient transfer learning strategies 
to effectively adapt large-scale VLMs to downstream tasks across diverse photographic image datasets~\cite{zhou2022coop,yao2023kgcoop,zhou2022cocoop,zhu2023prograd,cheng2023meta,gao2024clipadapter,yang2024mma,guo2025mmrl}.

\begin{figure*}[t]
    \vspace{-2.5em}
    \centering
    \begin{subfigure}[b]{0.46\textwidth}
        \includegraphics[width=0.97\textwidth]{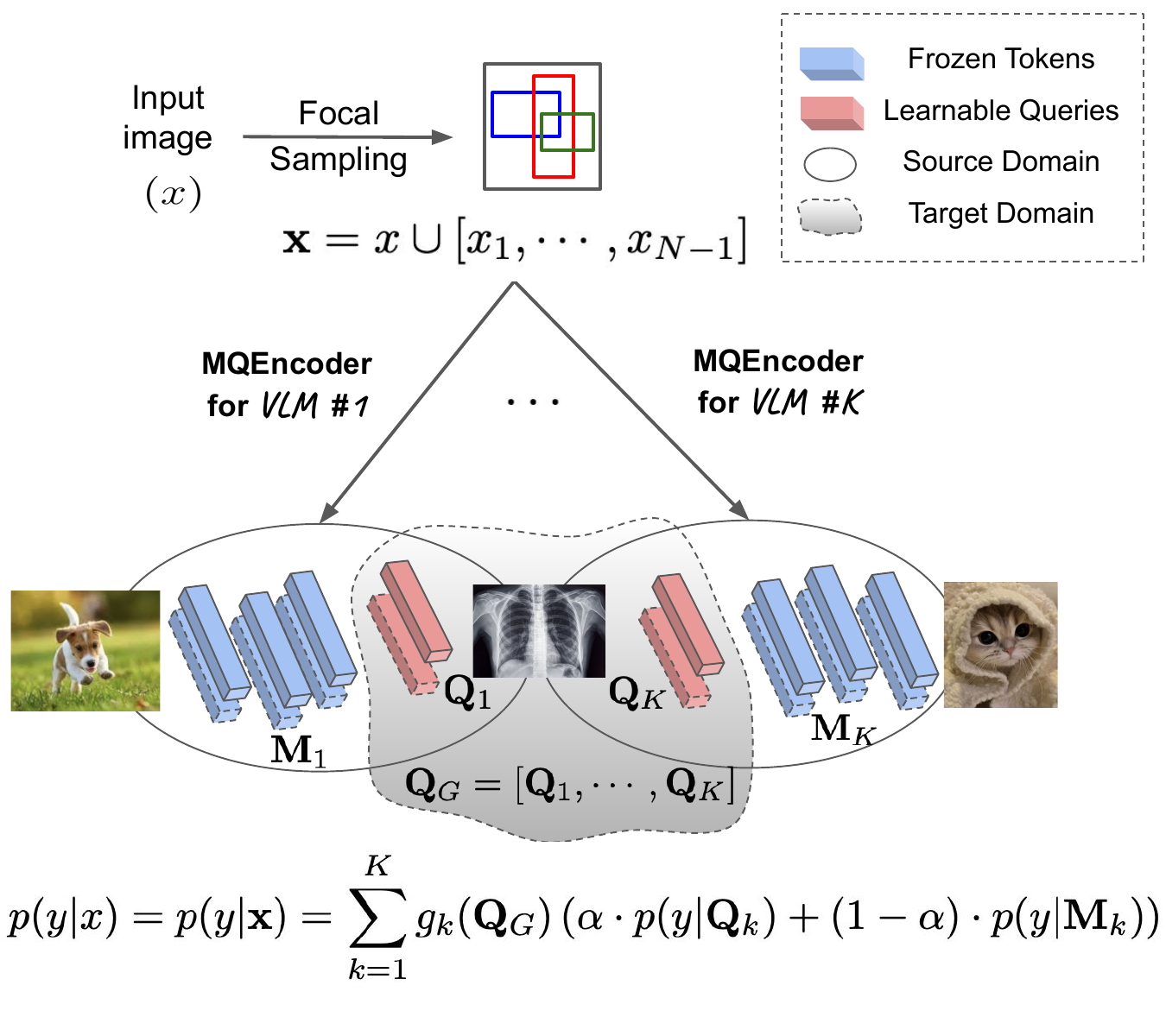}
        \caption{}
        \label{fig:demo_concept}
    \end{subfigure}
    \begin{subfigure}[b]{0.52\textwidth}
        \includegraphics[width=0.98\textwidth]{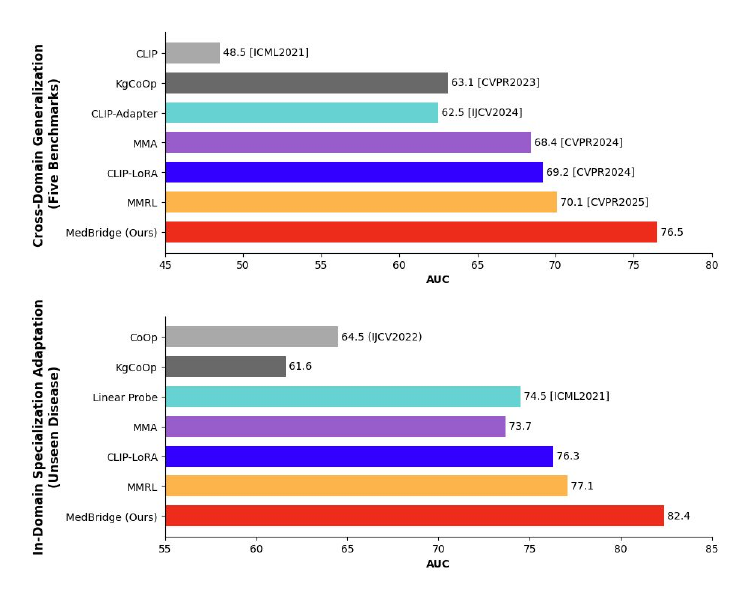}
        \caption{}
        \label{fig:demo_comp}
    \end{subfigure}
    \vspace{-0.8em}
      \caption{
      (a) To predict image labels $p(y|x)$, MedBridge first decomposes $x$ into $N$ local views $\mathbf{x}$, which are subsequently processed through a mixture of MQEncoders, each coupled with a VLM expert. Each expert outputs a set of frozen tokens \(\mathbf{M}\) and generates a small set of learnable multi-view queries \(\mathbf{Q}\) that interact with \(\mathbf{M}\).
      A gating network driven by the learnable queries 
      dynamically weights the experts, 
      and the final prediction combines \(\mathbf{M}\) and \(\mathbf{Q}\) via a factor \(\alpha\). The added learnable queries 
      increase only approximately 4\% more parameters compared to the base models, while effectively steering heterogeneous foundation models toward the target domain's representation. 
      (b) MedBridge yields significant improvements in disease classification across various medical benchmarks. 
      }
    \label{fig:demo}
    \vspace{-1.1em}
\end{figure*}

Despite recent advances, adapting foundation VLMs to \textit{medical imaging} remains challenging due to 
three simultaneous and mutually reinforcing constraints that existing adaptation methods do not address jointly. First, medical images are typically high-resolution ($\sim2048 \times 2048$) and contain sparse, spatially localized pathological evidence. Clinically relevant findings such as subtle nodules often vanish under the aggressive downsampling required by standard VLM input resolution (e.g., $224 \times 224$), leading to a critical loss of diagnostic information. Second, there exist significant domain discrepancies between natural and medical images, spanning visual statistics and label semantics. Adaptation strategies that directly perturb pretrained weights risk destroying the generalizable source knowledge that makes foundation models valuable, particularly in data-scarce clinical settings. Third, medical diagnosis is intrinsically multi-label and heterogeneous, where a single image often contains several co-existing disease labels indicated by distinct and often non-overlapping visual cues~\cite{holste2024towards} (e.g., cardiomegaly requires global structural understanding while pneumothorax hinges on fine local edge patterns). This diversity of diagnostic evidence imposes a requirement for multi-scale and multi-perspective reasoning, which is difficult for a single foundation model to capture effectively (\cref{fig:demo_comp}).
While recent efforts have leveraged foundation VLMs for preliminary medical tasks such as anomaly detection~\cite{huang2024adapting,zhang2024mediclip} and segmentation~\cite{poudel2024exploring,zhang2022adapting}, the challenge of achieving flexible and efficient adaptation for \textit{diagnosis}, a cornerstone of clinical practice, remains largely unaddressed. 
To jointly address these challenges, we introduce a new direction for domain adaptation, \emph{MedBridge}, that flexibly adapts arbitrary foundation VLMs to multi-label medical image diagnosis via query-mediated expert bridging.  
As shown in \cref{fig:demo_concept}, MedBridge transforms frozen foundation VLMs into multi-view query encoders augmented with a query-guided mixture-of-experts (MoE) routing mechanism. The multi-view design preserves the sparse high-resolution local evidence, and a set of learnable query tokens is injected directly into the vision encoder, serving as a unified adaptation interface that \textit{(i) bridges the domain gap via representation interaction, (ii) aggregates multi-view evidence, and (iii) drives multi-expert routing}. 
This design enables non-destructive and flexible adaptation: pretrained VLMs remain frozen to preserve generalizable knowledge, while adaptation is achieved through interaction with a compact set of learnable parameters. Meanwhile, the query-driven MoE dynamically integrates heterogeneous foundation models, capturing complementary diagnostic cues for multi-label reasoning. Together, it forms a synergistic framework that jointly addresses resolution constraints, domain shift, and diagnostic heterogeneity with minimal overhead.
As a result, MedBridge is model-agnostic and supports both cross-domain adaptation from various general-purpose VLMs, including CLIP~\cite{radford2021clip}, LLaVA~\cite{liu2023llava}, Qwen-VL~\cite{bai2025qwen3}, and further refinement of pre-trained medical ones like CheXzero~\cite{tiu2022chexzero}, MedGemma~\cite{sellergren2025medgemma}, maintaining strong performance even with limited training data. 
Across five benchmark medical datasets for multi-label thoracic disease classification, MedBridge consistently outperforms state-of-the-art adaptation methods, offering substantial advantages through improved accuracy, reduced computational costs, and robust performance under different adaptation scenarios (\cref{fig:demo_comp}). Our key contributions are:
\vspace{-0.8em}
\begin{itemize}
    \item \textbf{A query-driven adaptation interface for foundation VLMs.}
    We introduce multi-view query encoders (MQEncoders) that inject learnable queries into intermediate layers of frozen backbones, enabling non-destructive alignment while preserving fine-grained pathology.

    \vspace{-0.4em}
    \item \textbf{Flexible query-conditioned mixture-of-experts routing.}
    The learnable queries jointly serve as adaptation variables and routing signals, enabling dynamic architecture-agnostic integration of heterogeneous VLMs without requiring a shared representation space.

    \vspace{-0.4em}
    \item \textbf{Strong and generalizable empirical gains}
    of MedBridge across five medical datasets and three adaptation tasks, demonstrating robustness and model-agnostic transfer.
\end{itemize}

\vspace{-1em}
\section{Related Work}
\label{sec:related_work}
\vspace{-0.8em}

\textbf{Vision-language models} like CLIP~\cite{radford2021clip}, LLaVA~\cite{liu2023llava}, SigLIP~\cite{zhai2023siglip}, OpenCLIP~\cite{cherti2023openclip}, Gemini~\cite{team2023gemini}, and Gemma~\cite{team2024gemma} have made significant progress by leveraging large-scale training on massive datasets with billions of image-text pairs~\cite{schuhmann2022laion}, achieving strong performance across diverse downstream tasks. 
However, their direct deployment in the medical domain remains limited, e.g., CLIP in \cref{fig:demo_comp}, due to a substantial domain shift~\cite{tiu2022chexzero,shakeri2024few,zhao2025clipsurvey}, necessitating further domain-specific adaptation.

\vspace{-0.2em}
\noindent\textbf{Parameter-efficient transfer learning for VLMs.}
As VLMs scale, fine-tuning all model parameters is costly and prone to overfitting or catastrophic forgetting. Thus, recent work focused on parameter-efficient transfer learning, primarily through prompt learning and adapter modules. In prompt learning, CoOp~\cite{zhou2022coop} turns fixed templates to be learnable, CoCoOp~\cite{zhou2022cocoop} generates instance-specific prompts with visual features, ProGrad~\cite{zhu2023prograd} and KgCoOp~\cite{yao2023kgcoop} retain pre-trained knowledge during adaptation.
Adapter-based methods introduce lightweight modules: CLIP-Adapter~\cite{gao2024clipadapter} refines representations using MLPs, Tip-Adapter~\cite{zhang2022tipadapter} enables test-time inference via feature caching, MMA~\cite{yang2024mma} and MMRL~\cite{guo2025mmrl} incorporate a shared space for multimodal interaction.
CLIP-LoRA~\cite{zanella2024low} employs the low-rank adaptation.
These adaptation strategies are primarily tailored for in‑domain, cross‑dataset generalization and may not transfer well to cross‑domain multi-label medical data (\cref{fig:demo_comp}). In contrast, MedBridge goes beyond conventional adapters by explicitly addressing medical domain discrepancies, enabling robust domain adaptation from arbitrary foundation VLMs to medical data.

\vspace{-0.2em}
\noindent\textbf{Adapting foundation VLMs to the medical domain.}
Recent studies have extended general-purpose VLMs to medical report generation~\cite{pellegrini2025radialog,perez2025exploring},
anomaly detection~\cite{huang2024adapting,zhang2024mediclip}, segmentation~\cite{poudel2024exploring,zhang2022adapting}, 
reconstruction~\cite{chambon2022adaptingnipsw}, pathology imaging~\cite{lai2023clipath}, object detection~\cite{qin2022medical}.
However, to the best of our knowledge, a flexible adaptation of arbitrary foundation VLMs for multi-label disease diagnosis is largely underexplored. Current efforts mostly either carry out large-scale VLM pre-training to build medical-specific foundation models~\cite{zhou2024benchx} (GLORIA~\cite{huang2021gloria}, ConVIRT~\cite{zhang2022convirt}, MedCLIP~\cite{wang2022medclip}, CheXzero~\cite{tiu2022chexzero}, 
KAD~\cite{zhang2023kad}, MedKLIP~\cite{wu2023medklip}, MGCA~\cite{wang2022mgca}, MAVL~\cite{phan2024decomposing}, MedGemma~\cite{sellergren2025medgemma}) with substantial computational costs, or adapting VLMs that have been pre‑trained on medical data~\cite{shakeri2024few,zhang2024disease,lian2024less,jiang2024med}. 
However, we aim for a new direction of lightweight domain adaptation that rapidly re-purposes arbitrary foundation VLMs or refines pre-trained medical ones for unseen disease diagnosis, avoiding the heavy overhead of full re-training while maintaining strong clinical performance.

\vspace{-0.8em}
\section{Proposed Method}
\label{sec:method}
\vspace{-0.6em}

\begin{figure}[t]
  \centering
  \vspace{-1em}
  \includegraphics[width=0.85\linewidth]{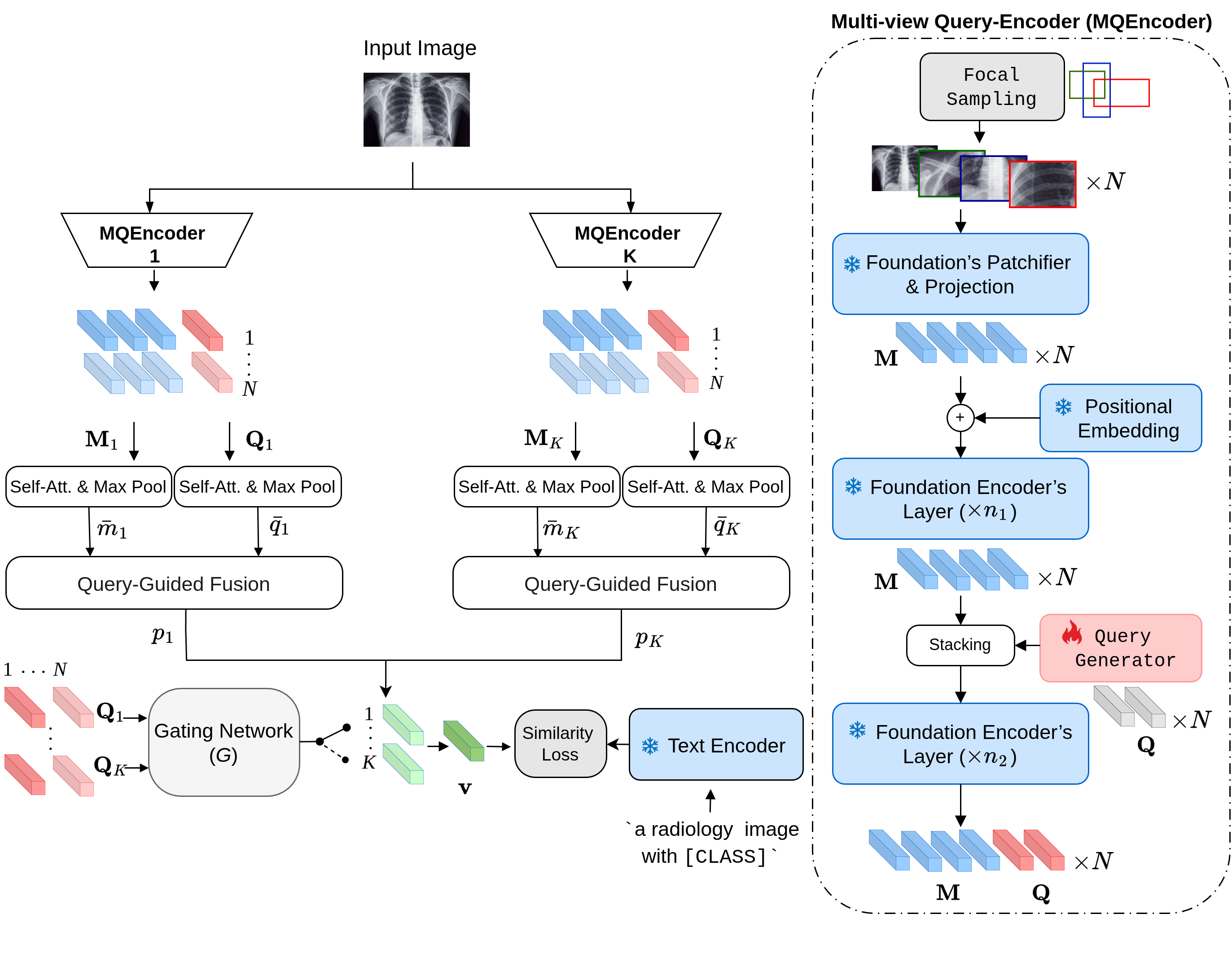}
  \vspace{-1.2em}
  \caption{\emph{MedBridge} framework: 
  MQEncoder extracts fine-grained regions from the high-resolution input image and encodes them into frozen tokens and learnable queries for lightweight adaptation. An MoE module routes these tokens through the most relevant encoders, and the final prediction token adds the query tokens to the frozen ones with a hyperparameter weight \(\alpha\).
}
  \label{fig:framework}
  \vspace{-1.1em}
\end{figure}

Let \(\mathcal{D}_s \) and \( \mathcal{D}_t \) denote the source and target domains respectively, and $K$ the number of ViT-based foundation VLMs that are trained in the source domain.
The goal is to adapt and specialize these general-purpose VLMs for multi-label disease diagnosis on medical data \(\mathcal{D}_t = \{ (x_t^{i}, y_t^{i}) \}_{i=1}^{N_t}\). Each image $x_t \in \mathbb{R}^{H\times W\times 3}$ is annotated with ground-truth labels
$y_t \in \{0,1\}^{L}$, indicating the presence of $L$ possible diseases: \(y_{t,j}\) is 1 if the \(j\)-th disease is present otherwise \(0\), allowing multiple diseases to be indicated simultaneously. 
We convert $y_t$ into a list of textual prompts in the template of `a radiology image with [CLASS]', and encode them using the corresponding text encoder from the model.
Our proposed MedBridge (\cref{fig:framework}) presents a new paradigm for domain adaptation with two key modules: 
I) Each MQEncoder wraps a foundation VLM encoder, extracts high-resolution sub-images from the input to capture fine-grained pathological cues, and encodes them into tokens using frozen backbone weights, while injecting a small set of learnable queries into intermediate layers to bridge the domain gap through direct interaction with frozen representations. These query tokens serve a dual role: adapting the frozen backbone to medical semantics, while producing the architecture-agnostic routing representations that enable the subsequent Mixture-of-Experts (MoE) integration.
II) An MoE routing building directly on the query-augmented tokens and dynamically routes across heterogeneous VLMs. 
Detailed steps are described below.

\noindent\textbf{Multi-view Query-Encoder (MQEncoder)}: 
{One image contains the richness of thousands}, especially in medical imaging, where disease manifestations can appear as multiple subtle regions, which may vanish when images are downsampled to the default input size expected by most foundation VLMs. To better preserve detailed pathological cues, MQEncoder first deploys a multi-view focal sampling strategy to augment the input with multiple fine-grained local views. 
From each high-resolution X-ray image, it extracts `$N-1$' local regions via either random selection or a sliding window screening through the full image with overlap, and stacks them alongside the resized full global image to create an $N$-view input $\mathbf{x}_t \in \mathbb{R}^{N\times H\times W\times 3}$. 
It enhances the visibility of disease-related regions and reduces the model's sensitivity to precise image alignments, ensuring more robust detection regardless of slight variations in positioning. It also acts as an implicit ensemble, expanding spatial coverage while mitigating overfitting and overdependence on specific image regions. 

Passing each view through the frozen vision encoder produces a sequence of $M$ tokens with embedding dimension $f$. We denote these tokens for view $v$ at layer $\ell$ as $\mathbf{M}^{(v,\ell)} \in \mathbb{R}^{M \times f}$. \emph{The MQEncoder keeps all $n$ layers of the foundation VLM encoder frozen}. Starting from layer `$n_1 + 1$', it introduces a set of $Q$ learnable query tokens $\mathbf{Q}^{(v,\ell)} \in \mathbb{R}^{Q \times f}$ (for $\ell > n_1$), which interact with the frozen tokens $\mathbf{M}$ to reduce the representation gap between source and target domains. The query tokens are initialized from a normal distribution $\mathcal{N}(0, 0.02)$ and shared across all views. For each view $v$, after layer $n_1$, we construct the concatenated token sequence:
\begin{equation}
\label{eq:n1}
\mathbf{X}^{(v,\ell)} = \big[\mathbf{M}^{(v,\ell)};\mathbf{Q}^{(v,\ell)}\big] \in \mathbb{R}^{(M+Q)\times f}.
\end{equation}
The remaining $n_2=n-n_1$ layers update $\mathbf{X}^{(v,\ell)}$ through attention mapping:
\begin{equation}
\label{eq:n2}
\begin{split}
&\mathbf{Z}^{(v,\ell)} = \mathbf{X}^{(v,\ell)} + \texttt{MHSA}(\texttt{Norm}(\mathbf{X}^{(v,\ell)})), \\
&\mathbf{X}^{(v,\ell+1)} = \mathbf{Z}^{(v,\ell)} + \texttt{FFN}(\texttt{Norm}(\mathbf{Z}^{(v,\ell)})),
\end{split}
\end{equation}
\noindent where $\ell=n_1+1,\cdots, n_1+n_2$; \texttt{MHSA} and \texttt{FFN} denote the multi-head self-attention and feed-forward network in the VLM backbones, respectively, and \texttt{Norm} the layer normalization~\cite{ba2016layernorm}. 
After the final layer, we split $\mathbf{X}^{(v,n)}$ back into the foundation and query tokens:
$\big[\mathbf{M}^{(v,n)};\mathbf{Q}^{(v,n)}\big] = \mathbf{X}^{(v,n)}$. 
It is worth noting that MQEncoder differs from QFormer~\cite{li2023blip}. In QFormer, the learnable queries first undergo self-attention and then interact with image features via cross-attention, which later interact with text via self-attention layers. 
In contrast, MQEncoder injects a compact set of learnable queries directly into the frozen vision encoder layers to effectively interact with the frozen tokens.
These auxiliary queries exploit contextual information from the source domain and are optimized for the downstream task, enabling them to learn reliable and domain-specific features. 

We then aggregate tokens across the $N$ views by applying self-attention to the concatenated multi-view tokens to extract their contextual relationships, and then a max-pooling to obtain one candidate token, $\bar{\mathbf m} \in \mathbb{R}^{f}$ and $\bar{\mathbf q} \in \mathbb{R}^{f}$ per VLM for the frozen and query tokens, respectively. 
Each expert outputs $\bar{\mathbf m}$ fused with $\bar{\mathbf q}$ (Query-Guided Fusion block), forming a prediction token $\mathbf{p}$:
\begin{equation}
\label{eq:alpha}
\mathbf{p}_k = \bar{\mathbf m}_k + \alpha \cdot \bar{\mathbf q}_k, \quad k=1,\cdots,K
\end{equation}
\noindent where $\alpha\geq 0$ is a positive weighting hyperparameter that controls the contribution of the query tokens to the final prediction ones. 
The impact of the weighting hyperparameter $\alpha$ on the model's behavior is summarized below and 
an ablation study is presented in \cref{sec:ablation}.
\vspace{-0.5em}
\begin{itemize}
    \item \textbf{$\alpha \approx 0$}: The prediction is predominantly determined by the frozen tokens of the foundation VLMs, resulting in a substantial domain gap.
    \vspace{-0.2em}
    \item \textbf{Intermediate values of $\alpha$}: As $\alpha$ increases, more weight is assigned to the query tokens, enabling domain-adaptive representations to contribute more to the final prediction and thereby reducing the domain discrepancy.
    \vspace{-0.2em}
    \item \textbf{$\alpha \geq 1$}: In this case, $\mathbf{p}_k$ becomes almost entirely dependent on the query tokens, leading to performance degradation (\cref{fig:input_crop}b): a limited set of query tokens cannot provide a sufficient representation of the input image.
\end{itemize}
\vspace{-0.2em}


\noindent\textbf{Mixture of Experts (MoE)}: Unlike the traditional MoE mechanism in VLMs, which uses all or part of tokens of the \emph{input} for routing, our MoE is driven by a small set of multi-view learnable queries. It selects the most relevant representations from weighted tokens (Eq.~\ref{eq:alpha}) while simultaneously updating query embeddings to address the domain gap. This query-driven routing facilitates a flexible integration of heterogeneous models, even those without a common representation space: capabilities that standard token-level MoE cannot provide. As shown in Figure~\ref{fig:framework}, the MoE takes the concatenated $N$-view learnable queries $\mathbf{Q}_k \in \mathbb{R}^{N\times Q\times f}, k=1,\dots,K$, from $K$ expert encoders as input to a gating network $G$, which dynamically computes and weights the contribution of each expert. 
Given an input token set $\mathbf{Q}_G=\{\mathbf{Q}_k\}_{k=1}^K$, the gating network computes a weight vector $\mathbf{g}(\mathbf{Q}_G) = [g_1(\mathbf{Q}_G), g_2(\mathbf{Q}_G), \ldots, g_K(\mathbf{Q}_G)]$, where each weight $g_k(\mathbf{Q}_G)$ is obtained via a softmax layer:
\vspace{-0.7em}
\begin{align}
g_k(\mathbf{Q}_G) &= \frac{\exp(\mathbf{w}_k^\top \mathbf{Q}_G + b_k)}{\sum_{j=1}^K \exp(\mathbf{w}_j^\top \mathbf{Q}_G + b_j)}, \quad
\sum_{k=1}^K g_k(\mathbf{Q}_G)= 1, \quad g_k(\mathbf{Q}_G) \geq 0.
\end{align} 
\noindent Here, $\mathbf{w}_k$ and $b_k$ are learnable parameters of the gating network for the $k$-th expert, reflecting the relevance of the $k$-th expert for the input. The final vision embedding $\mathbf{v}$ is a weighted sum of the prediction tokens from $K$ experts: 
\(
\mathbf{v} = \sum_{k=1}^K g_k(\mathbf{Q}_G) \cdot \mathbf{p}_k
\). 
In summary, the query generator offers a compact, plug-in interface that can be flexibly integrated into any foundation model, supporting the integration of diverse VLM experts. The MoE leverages these tokens to smooth the interaction between source and target domains, enabling a fast adaptation on the downstream diagnostic task.

\noindent\textbf{Text encoding:} Given the $L$ possible disease labels, each expressed as a textual prompt in the template of `a radiology image with [CLASS]', we tokenize them and pass them through the frozen text encoder of the foundation VLM. We take the embedding of the end-of-text (EOT) token from the final layer of the text encoder, and project it using a learnable projection layer to align with the vision embedding space, producing $L$ text features $\{\mathbf{z}_l\}^L_{l=1}$. 

\noindent\textbf{Loss criterion:} 
Let \(\mathbf{v}\) and $\{\mathbf{z}_l\}^L_{l=1}$ denote the  MoE and the textual label features of embedding size \(f\), respectively. We compute the cosine similarity between these via $\texttt{sim}(\mathbf{v}, \mathbf{z}_l) = \frac{\mathbf{v} \cdot \mathbf{z}_l}{\|\mathbf{v}\|{\|\mathbf{z}_l\|}}$, where $\|\cdot\|$ represents the $L_2$ norm. The resulting scores are treated as the predicted logits.
Since medical images can be associated with multiple disease labels, we optimize a multi-label binary cross-entropy (BCE) loss, pushing similarities for present disease labels toward 1 and for absent ones toward 0. We also ablated using the same BCE loss between the auxiliary clinical report features (when available) and the textual label features as additional supervision during adaptation in the appendix.











\vspace{-0.8em}
\section{Experimental Results}
\label{sec:results}
\vspace{-0.8em}

\noindent\textbf{Datasets}: We evaluated MedBridge and state-of-the-art adaptation approaches across five distinct medical imaging datasets: MIMIC-CXR v2~\cite{johnson2019mimic} (n = 227k), CheXpert Plus~\cite{chambon2024chexpertplus} (n = 224k), NIH ChestX-ray14~\cite{wang2017chestx} (n = 112k), RSNA Pneumonia~\cite{shih2019rsna} (n = 30k), and COVIDx CXR-4~\cite{wu2023covidx} (n = 84k). Among them, MIMIC-CXR, CheXpert Plus, and NIH ChestX-ray14 are multi-label datasets, where each subject may exhibit one or more of 14 common thoracic diseases. 
We evaluated the multi-label classification on all \emph{14} observations. In contrast, RSNA Pneumonia consists of binary labels that distinguish normal cases from pneumonia, while COVIDx CXR-4 includes cases labeled either as no findings or COVID-19. In addition, MIMIC-CXR and CheXpert Plus also provide paired radiology reports. Further details are provided in the appendix.
\begin{wrapfigure}{r}{0.5\textwidth}
\vspace{-1em}
    \centering
    \includegraphics[width=\linewidth]{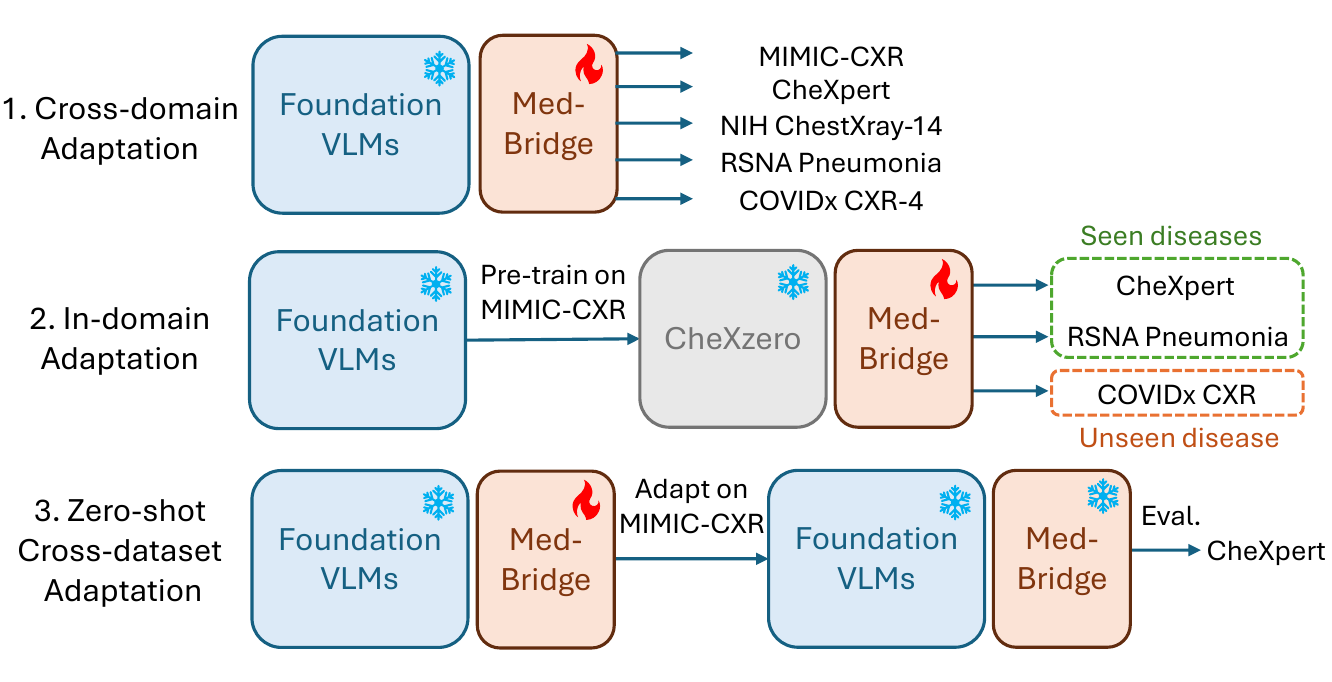}
    \vspace{-1.em}
    \caption{We evaluate MedBridge in three key adaptation tasks: cross-domain, in-domain, and zero-shot cross-dataset adaptation.}
    \label{fig:exp_types}
    \vspace{-1.2em}
\end{wrapfigure}
\noindent\textbf{Evaluation metrics}: 
We include standard medical image classification metrics: AUC, macro‑averaged F1 score, and accuracy (ACC). Following the best practice~\cite{seyyed2021underdiagnosis}, the binary decision threshold was chosen to maximize the F1 score, and ACC was also calculated under this threshold. All metrics are macro-averaged over all disease classes present in the target dataset, reported in percentage (\%). Computational cost comparisons are in the appendix. \\
\noindent\textbf{Baselines}: 
We compared state-of-the-art adaptation methods for VLMs, including zero-shot and full fine-tuning with CLIP~\cite{radford2021clip}, prompt learning: CoOp~\cite{zhou2022coop}, CoCoOp~\cite{zhou2022cocoop}, ProGrad~\cite{zhu2023prograd}, KgCoOp~\cite{yao2023kgcoop}, adapter-style learning: linear probing (LP)~\cite{radford2021clip}, CLIP-Adapter~\cite{gao2024clipadapter}, MMA~\cite{yang2024mma}, MMRL~\cite{guo2025mmrl}, a low-rank adapter CLIP-LoRA~\cite{zanella2024low}, and a medical VLMs adapter LP + text~\cite{shakeri2024few}. \\
\noindent\textbf{Implementation details}: 
We used the pre-trained CLIP~\cite{radford2021clip} with a ViT-B/16 backbone and 224$\times$224 input resolution, as the standard foundation VLM for adaptation, for a fair comparison with baselines. The default MoE setup integrates SigLIP~\cite{zhai2023siglip} and DINOv2~\cite{oquabdinov2} for efficiency. 
We further employed the pre-trained LLaVA-1.6 7B~\cite{liu2023llava}, Qwen3-VL 2B~\cite{bai2025qwen3}, and MedGemma-4B~\cite{sellergren2025medgemma} to evaluate the generalizability of MedBridge. 
The multi-view sampling extracted local regions at a size of 512$\times$512 from the full-resolution image and resized to 224$\times$224 as input, yielding $N=11$ input views obtained by a sliding window with overlap.
We used the AdamW optimizer at a learning rate of $5\times 10^{-4}$ and a batch size of 16, training for 3 epochs on a single NVIDIA H100 GPU.

\vspace{-0.8em}
\subsection{Evaluation Tasks and Results}
\vspace{-0.5em}

We conducted three key evaluations shown in~\cref{fig:exp_types} to assess MedBridge's adaptation capabilities: \\
\textbf{(1) Cross-domain adaptation} directly adapted general-purpose VLMs to medical data for diagnosis, simulating the challenging task of transferring knowledge from non-medical to medical domains. \\
\textbf{(2) In-domain adaptation} evaluated how well VLMs that are already pre-trained on large-scale medical data (e.g., CheXzero~\cite{tiu2022chexzero}), can adapt to new medical datasets, covering both previously seen and unseen disease scenarios. \\
\textbf{(3) Zero-shot cross-dataset evaluation} examined a domain generalization adaptation, where we first adapted general-purpose VLMs to the medical domain using a public medical dataset, then directly performed zero-shot multi-label classification on a different dataset without further fine-tuning.

Both cross- and in-domain adaptation were additionally evaluated under varying amounts of labeled training data (1\%, 10\%, 50\%, 100\%) to simulate real-world clinical scenarios with different data availability. Unless stated otherwise, the reported results of MedBridge are without clinical reports or MoE during adaptation to ensure a fair comparison with baselines.
We further evaluated MedBridge across different foundation VLMs to validate its generalizability in adapting arbitrary models.

\begin{table*}[t]
  \centering
  \scriptsize
  \caption{
  Evaluation of \emph{cross-domain adaptation} on five medical datasets test sets.
  }
  \label{tab:fine_tune}
  \setlength{\tabcolsep}{1.pt}
  \renewcommand{\arraystretch}{0.95}
  \begin{tabular}{l l*{15}{c}}
    \toprule
    \multirow{3}{*}{Method} & 
      \multicolumn{3}{c}{CheXpert Plus} &
      \multicolumn{3}{c}{MIMIC-CXR} &
      \multicolumn{3}{c}{NIH CXR-14} &
      \multicolumn{3}{c}{RSNA Pneu.} &
      \multicolumn{3}{c}{COVIDx CXR-4} \\
    \cmidrule(lr){2-4}\cmidrule(lr){5-7}\cmidrule(lr){8-10}\cmidrule(lr){11-13}\cmidrule(lr){14-16}
        & AUC~$\uparrow$ & F1~$\uparrow$ & ACC~$\uparrow$
        & AUC~$\uparrow$ & F1~$\uparrow$ & ACC~$\uparrow$
        & AUC~$\uparrow$ & F1~$\uparrow$ & ACC~$\uparrow$
        & AUC~$\uparrow$ & F1~$\uparrow$ & ACC~$\uparrow$
        & AUC~$\uparrow$ & F1~$\uparrow$ & ACC~$\uparrow$ \\
    \midrule
    CLIP (zero-shot)\conf{ICML2021}  & 51.07 & 28.85 & 19.65 & 51.74 & 22.40 & 13.46 & 50.11 & 16.30 & 9.65 & 38.98 & 37.37 & 22.98 & 50.74 & 66.67 & 50.00 \\
    CLIP (fine-tune)\conf{ICML2021}  & 68.47 & 40.16 & 74.43 & 58.09 & 26.64 & 55.73 & 54.51 & 17.78 & 40.09 & 73.17 & 48.41 & 65.89 & 59.60 & 69.41 & 57.98 \\
    CoOp\conf{IJCV2022}              & 66.66 & 36.72 & 68.16 & 63.69 & 28.57 & 70.96 & 57.98 & 17.78 & 61.43 & 82.10 & 58.38 & 76.91 & 56.89 & 66.77 & 50.54 \\
    CoCoOp\conf{CVPR2022}            & 61.50 & 35.03 & 65.62 & 59.86 & 26.97 & 54.80 & 52.12 & 16.68 & 36.87 & 81.58 & 57.88 & 76.05 & 56.16 & 66.67 & 50.01 \\
    ProGrad\conf{ICCV2023}           & 61.93 & 35.97 & 62.02 & 58.91 & 26.71 & 53.51 & 52.89 & 16.94 & 38.35 & 81.23 & 58.14 & 77.55 & 54.18 & 66.79 & 50.67 \\
    KgCoOp\conf{CVPR2023}            & 64.29 & 35.35 & 64.51 & 60.55 & 27.33 & 64.53 & 55.08 & 17.21 & 48.76 & 77.43 & 54.65 & 72.94 & 57.10 & 66.83 & 50.85 \\
    Linear Probing\conf{ICML2021}    & 68.45 & 37.69 & 71.48 & 65.20 & 29.13 & 68.87 & 58.22 & 18.45 & 57.59 & 81.39 & 58.00 & 77.96 & 58.31 & 66.91 & 51.27 \\
    LP + text\conf{MICCAI2024}       & 51.87 & 28.85 & 19.65 & 53.28 & 22.40 & 13.46 & 56.65 & 16.30 & 32.65 & 50.00 & 37.37 & 77.02 & 55.31 & 66.62 & 50.00 \\
    CLIP-Adapter\conf{IJCV2024}      & 62.24 & 35.11 & 60.52 & 57.20 & 25.93 & 49.63 & 51.00 & 16.68 & 33.45 & 78.46 & 55.08 & 73.16 & 58.87 & 66.69 & 50.09 \\
    CLIP-LoRA\conf{CVPRW2024}        & 75.91 & 47.88 & 82.13 & 66.30 & 30.62 & 72.05 & 59.90 & 19.54 & 50.17 & 84.63 & 60.95 & 79.01 & 62.15 & 67.69 & 53.53 \\
    MMA\conf{CVPR2024}               & 74.76 & 42.98 & 80.76 & 64.03 & 30.16 & 65.44 & 57.83 & 18.63 & 47.99 & 81.86 & 57.30 & 76.76 & 65.06 & 67.62 & 53.20 \\
    MMRL\conf{CVPR2025}              & 76.46 & 44.64 & 80.68 & 64.74 & 29.43 & 65.03 & 60.79 & 19.42 & 66.71 & 82.14 & 57.67 & 76.01 & 66.90 & 67.92 & 55.88 \\
    \midrule
    MedBridge (ours) & \underline{81.49} & \underline{47.30} & \textbf{85.01} & \underline{70.59} & \underline{33.04} & \underline{76.41} & \underline{65.03} & \underline{20.86} & \textbf{69.70} & \underline{85.38} & \underline{62.08} & \underline{79.35} & \underline{73.81} & \underline{73.22} & \underline{67.44} \\
    MedBridge (ours, w/ MoE)   & \textbf{83.55} & \textbf{49.94} & \underline{83.62} & \textbf{71.92} & \textbf{33.95} & \textbf{77.84} & \textbf{66.42} & \textbf{22.36} & \underline{69.58} & \textbf{86.26} & \textbf{62.97} & \textbf{79.50}  & \textbf{76.37} & \textbf{75.47} & \textbf{70.51}  \\
    \bottomrule
  \end{tabular}
  \vspace{-1.5em}
\end{table*}

\begin{table*}[h]
\scriptsize
  \caption{AUC (\%) of \emph{cross-domain adaptation} on test sets under different training data portions.}
  \vspace{-0.5em}
  \label{tab:cross_data_efficiency}
  \centering
  \setlength{\tabcolsep}{4.pt}
  \renewcommand{\arraystretch}{0.95}
  \begin{tabular}{lcccc|cccc|cccc}
    \toprule
    Task & \multicolumn{12}{c}{Cross-Domain Adaptation} \\\cmidrule(lr){2-13}
    \multicolumn{1}{l}{Dataset} &
    \multicolumn{4}{c}{CheXpert Plus} &
    \multicolumn{4}{c}{RSNA Pneumonia} &
    \multicolumn{4}{c}{COVIDx CXR-4} \\ \cmidrule(lr){2-5}\cmidrule(lr){6-9}\cmidrule(lr){10-13}
    Data Portion & 1\% & 10\% & 50\% & 100\% & 1\% & 10\% & 50\% & 100\% & 1\% & 10\% & 50\% & 100\% \\ \midrule
    CoOp\conf{IJCV2022}  & 56.79 & 65.83 & 69.95 & 66.66 & \underline{74.70} & \underline{79.04} & \underline{81.41} & 82.10 & 36.15 & 56.44 & 58.02 & 56.89 \\
    KgCoOp\conf{CVPR2023}  & \underline{58.42} & 57.62 & 57.78 & 64.29 & 62.86 & 67.74 & 69.29 & 77.43 & 38.44 & 53.29 & 54.99 & 57.10 \\
    LP\conf{ICML2021}  & 58.22 & 66.30 & 68.98 & 68.45 & 73.01 & 78.14 & 80.60 & 81.39 & \underline{62.63} & 59.09 & 59.51 & 58.31 \\
    CLIP-LoRA\conf{CVPRW2024} & 49.23 & 60.63 & 72.82 & 75.91 & 62.20 & 69.56 & 80.21 & \underline{84.63} & 61.87 & \underline{61.46} & 63.68 & 62.15
    \\
    MMA\conf{CVPR2024}  & 49.58 & \underline{68.96} & 76.33 & 74.76 & 33.05 & 58.08 & 81.29 & 81.86 & 55.37 & 61.35 & \underline{71.68} & 65.06  \\ 
    MMRL\conf{CVPR2025} & 58.71 & 66.80 & \underline{76.74} &  \underline{76.46} & 69.45 & 74.89 & 80.03 & 82.14 & 58.32 & 60.59 & 66.47 & \underline{66.90}  \\
    \midrule
    MedBridge (ours) & \textbf{64.52} & \textbf{69.64} & \textbf{77.22} & \textbf{81.49} & \textbf{76.27} & \textbf{79.84} & \textbf{84.16} & \textbf{85.38} & \textbf{67.75} & \textbf{74.42} & \textbf{76.14} & \textbf{73.81} \\ \bottomrule
  \end{tabular}
  \vspace{-1.8em}
\end{table*}

\noindent\textbf{Cross-Domain Adaptation:}
Here, we used frozen CLIP models pre-trained on large-scale natural image-text pairs as representative foundation VLMs, and adapted them with MedBridge and state-of-the-art multimodal adaptation approaches on multi-label thoracic disease classification across five medical datasets. \cref{sec::expert_vlms} reports the performance of MedBridge using other foundation VLMs. 
As shown in \cref{tab:fine_tune}, zero-shot classification using the pre-trained CLIP achieved only around 51\% AUC on average, highlighting the significant domain gap between natural and medical images and the need for effective adaptation. 
Fully fine-tuning CLIP on the respective datasets resulted in an average performance of about 62\%, a relatively limited gain that is likely attributed to overfitting.
Existing adaptation methods substantially improved performance, with CLIP-LoRA achieving an AUC of 75.91\% and MMRL achieving 76.46\% on CheXpert Plus. MedBridge further increased accuracy, outperforming MMRL by more than 7\% on the CheXpert Plus, MIMIC-CXR, and COVIDx CXR-4 datasets, and by approximately 5\% on NIH and RSNA Pneumonia, demonstrating its superior generalization and diagnostic performance.
The addition of the MoE module to MedBridge further improved its diagnostic precision, especially in CheXpert Plus and COVIDx CXR-4, increasing AUC by 2\% and 2.5\%, respectively. 
In addition, as shown in~\cref{tab:cross_data_efficiency}, MedBridge maintained consistently high performance across different training data proportions, highlighting its strong data efficiency.

\begin{wraptable}{r}{0.45\textwidth}
\vspace{-1em}
\scriptsize
  \caption{AUC scores (\%) of \emph{in-domain adaptation} with CheXzero on the test sets under different training data portions.}
  \vspace{-0.2em}
  \label{tab:in_data_efficiency}
  \centering
  \setlength{\tabcolsep}{2pt}
  \renewcommand{\arraystretch}{0.95}
  \begin{tabular}{lcccc|cccc}
    \toprule
    \multicolumn{1}{l}{Dataset} &
    \multicolumn{4}{c}{CheXpert Plus} &
    \multicolumn{4}{c}{COVIDx CXR-4} \\ 
    &
    \multicolumn{4}{c}{(Seen Disease)} &
    \multicolumn{4}{c}{(Unseen Disease)} \\ 
    \cmidrule(lr){2-5}\cmidrule(lr){6-9}
    Data Portion & 1\% & 10\% & 50\% & 100\% & 1\% & 10\% & 50\% & 100\% \\ \midrule
    CoOp & 58.5 & 64.9 & 70.3 & 72.1 & 50.6 & 55.4 & 57.6 & 56.6 \\
    KgCoOp  & 59.9 & 64.1 & 64.9 & 64.5 & 42.9 & 53.8 & 54.3 & 58.6  \\
    LP & \underline{64.3} & 64.8 & 74.2 & \underline{83.3}  & 60.1 & 61.6 & 65.5 & 65.7 \\
    CLIP-LoRA & 58.2 & 65.0 & \underline{77.8} & 81.5  & 52.3 & 58.6 & \underline{71.1} & 71.0
    \\
    MMA   & 49.2 & \textbf{73.6} & 77.7 & 77.9 & 50.9 & 60.8 & 65.2 & 69.4 \\ 
    MMRL & 59.4 & 69.6 & 76.5 & 80.7 & \underline{62.6} &  \underline{66.5} & 67.1 & \underline{73.4} \\
    \midrule
    MedBridge & \textbf{68.1} & \underline{71.5} & \textbf{78.3} & \textbf{86.6} & \textbf{72.9} & \textbf{73.6} & \textbf{75.2} & \textbf{78.1} \\ \bottomrule
  \end{tabular}
  \vspace{-1.8em}
\end{wraptable}
\noindent\textbf{In-Domain Adaptation:}
We also assessed the ability of MedBridge and other adaptation methods to adapt in-domain pre-trained VLMs, i.e., VLMs that have already been further pre-trained on medical datasets, to new medical datasets for diagnosing both previously seen and unseen diseases. To this end, we used CheXzero~\cite{tiu2022chexzero}, a self-supervised CLIP-based model pre-trained on the MIMIC-CXR dataset~\cite{johnson2019mimic}. 
Since the diseases in CheXpert Plus are also present in MIMIC-CXR, we used them to assess in-domain adaptation on seen diseases. In contrast, COVID-19 represents an unseen disease for the CheXzero model, thus, we utilized the COVIDx CXR-4 dataset to evaluate in-domain adaptation to unseen disease. We also investigated performance under varying amounts of labeled training data. 
\cref{tab:in_data_efficiency} reported the performance of MedBridge and six best-performed adaptation baselines. In-domain adaptation achieved an overall higher diagnostic accuracy compared to cross-domain adaptation due to the reduced domain gap. 
Existing adaptation methods performed well for the in-domain adaptation with seen diseases, i.e., CheXpert, as they were primarily designed for in-domain use.
MedBridge achieved comparably high performance on the seen diseases. 
Notably, in the case of the unseen disease (COVID-19), MedBridge achieved a significantly higher performance, with an AUC score of 78.1\%, outperforming the second-best adaptation method by 5\% while maintaining stable results across varying training data proportions. Given the common data scarcity in the medical domain, this setting reflects real-world needs for rapid adaptation to emerging diseases, highlighting the efficiency and practical utility of MedBridge in clinical applications.

\noindent\textbf{Zero-Shot Cross-Dataset Evaluation:}
\begin{table*}[b]
\vspace{-1.5em}
  \centering
  \scriptsize
  \caption{Results on zero-shot cross-dataset evaluation.}
  \vspace{-1em}
  \label{tab:zero_shot}
  \setlength{\tabcolsep}{3pt}
  \renewcommand{\arraystretch}{0.84}
  \begin{tabular}{c ccccccccccc}
    \toprule
    \multirow{2}{*}{Dataset} &  & \multicolumn{10}{c}{Method} \\
    \cmidrule{3-12}
    & & CoOp 
    & CoCoOp 
    & ProGrad 
    & KgCoOp 
    & LP 
    & CLIP-Ad. 
    & MMA 
    & MMRL 
    & MedBridge 
    & MedB. (MoE) \\
    \midrule
    \multirow{1}{*}{Source (MIMIC)} 
      & AUC~$\uparrow$
      & 63.69 & 59.86 & 58.91 & 60.55 & 65.20 & 57.20 & 64.03 & 64.74 & 70.59 & \textbf{71.92} \\
    \midrule
    \multirow{3}{*}{Target (CheXpert Plus)} 
      & AUC~$\uparrow$
      & 69.34 & 64.81 & 60.06 & 65.29 & 73.66 & 63.92 & 76.07 & 76.48 & 78.69 & \textbf{79.94} \\
      & F1~$\uparrow$
      & 38.72 & 37.76 & 34.14 & 36.55 & 40.55 & 36.50 & 46.84 & 45.20 & 45.01 & \textbf{47.15} \\
      & ACC~$\uparrow$
      & 77.15 & 70.27 & 62.80 & 72.46 & 76.31 & 71.10 & 83.78 & 82.38 & 84.78 & \textbf{85.12} \\
    \bottomrule
  \end{tabular}
  \vspace{-2.2em}
\end{table*}
We evaluated MedBridge for this task by first adapting the foundation VLM to the medical domain using the large-scale MIMIC-CXR dataset, aiming to narrow the domain gap between natural and medical images. We then conducted zero-shot multi-label diagnosis on the CheXpert Plus dataset, without further fine-tuning. As shown in \cref{tab:zero_shot}, MedBridge substantially outperformed state-of-the-art adaptation methods, achieving the highest AUC of 79.94\%, nearly 3.5\% higher than the next best method, MMRL. This highlights MedBridge's ability to enhance generalization across diverse medical datasets, allowing foundation VLMs fine-tuned on domain-specific public data to be reliably transferred to data‑limited local clinical settings.

\vspace{-1em}
\subsection{Adaptation of Various Foundation VLMs}
\vspace{-0.5em}
\begin{table}[t]
  \centering
  \vspace{-1.em}
  \scriptsize
  \caption{Evaluation of Medbridge with different foundation VLMs on three medical \emph{test} sets. Complete results are reported in the appendix.} 
  \label{tab:vlm_expert}
   \setlength{\tabcolsep}{4.8pt}
   \renewcommand{\arraystretch}{0.98}
  \begin{tabular}{l{l}l*{8}{c}}
    \toprule
    \multirow{2}{*}{Adaptation} & \multirow{2}{*}{Method} & 
      \multicolumn{3}{c}{CheXpert Plus} &
      \multicolumn{3}{c}{MIMIC-CXR} &
      \multicolumn{3}{c}{RSNA Pneumonia} \\ 
    \cmidrule(lr){3-5}\cmidrule(lr){6-8} \cmidrule(lr){9-11}
        && AUC~$\uparrow$ & F1~$\uparrow$ & ACC~$\uparrow$
        & AUC~$\uparrow$ & F1~$\uparrow$ & ACC~$\uparrow$ 
        & AUC~$\uparrow$ & F1~$\uparrow$ & ACC~$\uparrow$ \\
    \midrule
     \multirow{6}{*}{Cross-domain}& CLIP & 51.07 & 28.85 & 19.65 & 51.74 & 22.40 & 13.46 & 38.98    &   37.37    &   22.98 \\
     & LLaVA & 52.07 & 31.50 & 46.80 & 48.14 & 22.64 & 27.40 & 48.72 & 37.32 & 22.94 \\
     & Qwen3-VL & 66.86 & 39.68 & 72.90 & 59.33 & 26.73 & 64.31 & 75.37 & 53.81 & 77.29 \\
     & MedBridge (CLIP)     & 81.49 & 47.30 & 85.01 & 70.59 & 33.04 & 76.41 & 85.38 & 62.08 & 79.35 \\
    & MedBridge (LLaVA)    & \textbf{84.49} & \textbf{50.41} & 84.65 & \textbf{73.57} & \textbf{35.56} & 77.97 & \textbf{86.49} & \textbf{64.72} & 80.55 \\
    & MedBridge (Qwen3-VL) & 81.37 & 47.65 & \textbf{85.52} & 73.03 & 34.50 & \textbf{78.23} & 86.16 & 63.65 & \textbf{81.33} \\
    \midrule
    \multirow{2}{*}{In-domain} &
    MedGemma & 52.66 & 30.37 & 48.97 & 53.75 & 25.09 & 38.18 & 59.40 & 38.81 & 39.13  \\
    &MedBridge (MedGemma) & \textbf{85.83} & \textbf{56.06} & \textbf{85.34} & \textbf{76.88} & \textbf{38.19} & \textbf{81.69} & \textbf{89.68} & \textbf{70.11} & \textbf{85.46} \\
    \bottomrule
  \end{tabular}
  \vspace{-1em}
\end{table}
\label{sec::expert_vlms} To ensure a fair comparison with baseline methods, we used CLIP as the foundation VLM in the previous sections. However, MedBridge is highly versatile and designed to be VLM-agnostic. It can readily adapt different foundation VLMs with their respective visual and text encoders, such as MedGemma~\cite{sellergren2025medgemma} for in-domain adaptation, and LLaVA~\cite{liu2023llava}, Qwen3-VL~\cite{bai2025qwen3} for cross-domain adaptation. 
Although the text encoders from these models are not CLIP-style, they are used solely for label prompt encoding, while MedBridge primarily adapts the visual encoders.
As shown in \cref{tab:vlm_expert}, MedBridge successfully adapted these VLM backbones across CheXpert Plus, MIMIC-CXR, and RSNA Pneumonia datasets, outperforming their original performance by a large margin. We observed sequential performance gains, starting from CLIP (150M) and increasing with the larger Qwen3-VL (2B), LLaVA (7B) foundation model. The best results were achieved with the domain-specific MedGemma backbone, yielding the highest AUC of 85.83\% in CheXpert Plus, 76.88\% in MIMIC-CXR, and 89.68\% in RSNA Pneumonia.
It demonstrates the effectiveness and generalizability of MedBridge across diverse foundation VLMs.


\begin{wraptable}{r}{0.5\textwidth}
\vspace{-1.2em}
\centering
\scriptsize
\caption{Comparison of LoRA-based adaptation within the MedBridge framework.}
\vspace{-1em}
\setlength{\tabcolsep}{1.8pt}
\renewcommand{\arraystretch}{0.96}
\begin{tabular}{lcccccc}
\toprule
\multirow{2}{*}{Method} & \multicolumn{3}{c}{CheXpert} & \multicolumn{3}{c}{MIMIC} \\
\cmidrule(lr){2-4} \cmidrule(lr){5-7}
 & AUC & F1 & ACC & AUC & F1 & ACC \\
\midrule
CLIP-LoRA~\cite{zanella2024low} & 75.91 & 47.88 & 82.13 & 66.30 & 30.62 & 72.05 \\
MedBridge$_\text{LoRA}$ (w/ MoE) & 77.39 & 48.33 & 83.22 & 69.82 & 33.09 & 78.01 \\
MedBridge (w/ MoE) & 83.55 & 49.94 & 83.62 & 71.92 & 33.95 & 77.84 \\
\bottomrule
\end{tabular}
\vspace{-1.em}
\label{tab:lora_ablation}
\end{wraptable}
\vspace{-0.8em}
\subsection{LoRA as a drop-in replacement of Query Encoder}
\vspace{-0.5em}
We replace MQEncoder’s learnable query injection with LoRA while keeping the rest unchanged, denoted as MedBridge$_\text{LoRA}$. As shown in \cref{tab:lora_ablation}, MedBridge$_\text{LoRA}$ already improves over CLIP-LoRA~\cite{zanella2024low} (+1.48 AUC on CheXpert, +3.52 on MIMIC), highlighting gains from multi-view sampling and MoE. The full MedBridge further outperforms MedBridge$_\text{LoRA}$ (+6.16 AUC on CheXpert, +2.10 on MIMIC), isolating the benefit of learnable query injection. Unlike LoRA, which perturbs pretrained weights via low-rank updates, MQEncoder keeps pretrained representations frozen and adapts via flexible query interactions, which are more robust under domain shift and expert integration. It is also more computational efficient (peak memory: 3.9 vs. 10.3 GB, training time: 1219 vs. 1989 ms/step). 

\vspace{-0.8em}
\subsection{Generalization to Other Imaging Modality and Medical Application}
\vspace{-0.5em}

\begin{table}[b]
\vspace{-2.em}
\scriptsize
\centering
\caption{Generalization to the multi-label ODIR-5K retinal fundus imaging dataset.}
\label{tab:odir}
\setlength{\tabcolsep}{4.5pt}
\renewcommand{\arraystretch}{0.86}
\begin{tabular}{lccccccc}
\toprule
Metric 
& CLIP~\cite{radford2021clip} 
& ProGrad~\cite{zhu2023prograd} 
& KgCoOp~\cite{yao2023kgcoop} 
& CLIP-Ad.~\cite{gao2024clipadapter} 
& MMA~\cite{yang2024mma} 
& MMRL~\cite{guo2025mmrl} 
& MedBridge (Ours) \\
\midrule
AUC~$\uparrow$ & 44.84 & 60.28 & 67.08 & 56.36 & 61.35 & 69.52 & \textbf{72.58} \\
F1~$\uparrow$  & 23.58 & 32.55 & 40.25 & 29.39 & 31.99 & 40.40 & \textbf{45.45} \\
ACC~$\uparrow$ & 14.63 & 59.91 & 73.56 & 60.26 & 67.76 & 74.55 & \textbf{75.29} \\
\bottomrule
\end{tabular}
\vspace{-1.8em}
\end{table}

To assess the generalizability of MedBridge beyond chest radiography, we extend our evaluation to retinal fundus imaging using ODIR-5K dataset~\cite{shanggong2019odir5k}, a multi-label 8-class ocular disease classification benchmark. Crucially, we retain the same hyperparameters used for the chest radiography without any modality-specific tuning, providing a rigorous test of out-of-domain transferability.
Table~\ref{tab:odir} shows that MedBridge kept outperforming all baselines in this task, despite the large domain gap between color fundus photography and grayscale chest radiography. It indicates MedBridge is not tailored to a single modality but yields robust performance across heterogeneous medical modalities and tasks.


\vspace{-0.8em}
\subsection{Ablation Study}
\label{sec:ablation}
\vspace{-0.5em}

We performed comprehensive ablation studies to evaluate key designs in MedBridge,
reporting the average AUC on the \emph{validation sets} of three representative datasets: CheXpert Plus, MIMIC-CXR, and RSNA Pneumonia, with detailed and additional ablation results provided in the appendix.
\begin{wraptable}{r}{0.45\textwidth}
\centering
\vspace{-0.5em}
\scriptsize
\caption{Ablation on (upper) experts in MoE routing and (lower) removing key components of MedBridge on the \underline{validation} sets. Detailed results are reported in the appendix.}
\vspace{-0.2em}
\setlength{\tabcolsep}{1.pt}
\renewcommand{\arraystretch}{0.85}
\begin{tabular}{l|cc|cc}
\toprule
\multirow{2}{*}{Experts} & \multicolumn{2}{c|}{RSNA Pneu.}  & \multicolumn{2}{c}{CheXpert Plus}\\
    & AUC~$\uparrow$ & ACC~$\uparrow$ & AUC~$\uparrow$ & ACC~$\uparrow$ \\\midrule
    CLIP & 85.08 & 78.31 & 75.41 & 75.64  \\
    CLIP + SigLIP & 85.98 & 79.20 & 76.25 & 77.48  \\
    CLIP + SigLIP + DINOv2 & 86.26 & \textbf{82.84} & 76.44 & 76.27 \\
    CLIP + EVA-CLIP & 86.66 &  82.13 & 76.87 & 77.86 \\
    CLIP + LLaVA & 86.69 & 82.02 & \textbf{77.97} & \textbf{79.28} \\  
    CLIP + LLaVA + MetaCLIP & \textbf{86.95} & 81.60 & 77.56 & 78.22 \\   
\midrule    
    CLIP + MedGemma & 88.50 & 83.06 & 78.61 & 77.23 \\   
    CLIP + LLaVA + MedGemma & \textbf{89.26} & \textbf{83.48} & \textbf{80.65} &  \textbf{79.95} \\   
\bottomrule
\end{tabular}
\renewcommand{\arraystretch}{0.85}
\begin{tabular}{l|cc|cc}
\toprule
& \multicolumn{2}{c|}{RSNA Pneu.}  & \multicolumn{2}{c}{CheXpert Plus}\\
Method & AUC $\uparrow$ & ACC $\uparrow$ & AUC $\uparrow$ & ACC $\uparrow$ \\
\midrule
MedB. (w/o learnable queries) & 81.81 & 73.74 & 72.56 & 72.77 \\
MedB. (w/o multi-view sampl.) & 85.50 & 78.76 & 77.28 & 77.97 \\
MedB. (w/o MoE)                & 85.08 & 78.31 & 75.41 & 75.64 \\
\midrule 
MedBridge (full)   & 89.26 & 83.48 & 80.65 & 79.95 \\
\bottomrule
\end{tabular}
\label{tab:moe}
\vspace{-1.5em}
\end{wraptable}
\noindent \textbf{MoE Module}: 
\cref{tab:moe} (upper) shows the impact of MoE with different foundation models. While using CLIP alone provided a baseline with an AUC of 85.08\% in RSNA Pneumonia and 75.41\% in CheXpert Plus, the inclusion of additional diverse and more specialized experts substantially improved performance.
Although experts such as DINOv2 and LLaVA are not trained with CLIP-style alignment, MedBridge's flexible adaptation strategy leverages learnable query tokens to effectively integrate their representations in a synergistic manner, resulting in enhanced diagnostic capability.
In RSNA, AUC improved to 86.95\% in the cross-domain setting (CLIP+LLaVA+MetaCLIP) and further to 89.14\% in the in-domain setting (CLIP+LLaVA+MedGemma), and results on CheXpert Plus increased significantly to an AUC of 80.65\%.
Detailed results in the appendix 
further illustrate that the diversity of expert architectures and training domains matters more than the number of experts. 
\begin{wrapfigure}{r}{0.38\textwidth}
    \centering
    \vspace{-4.em}
\includegraphics[width=\linewidth]{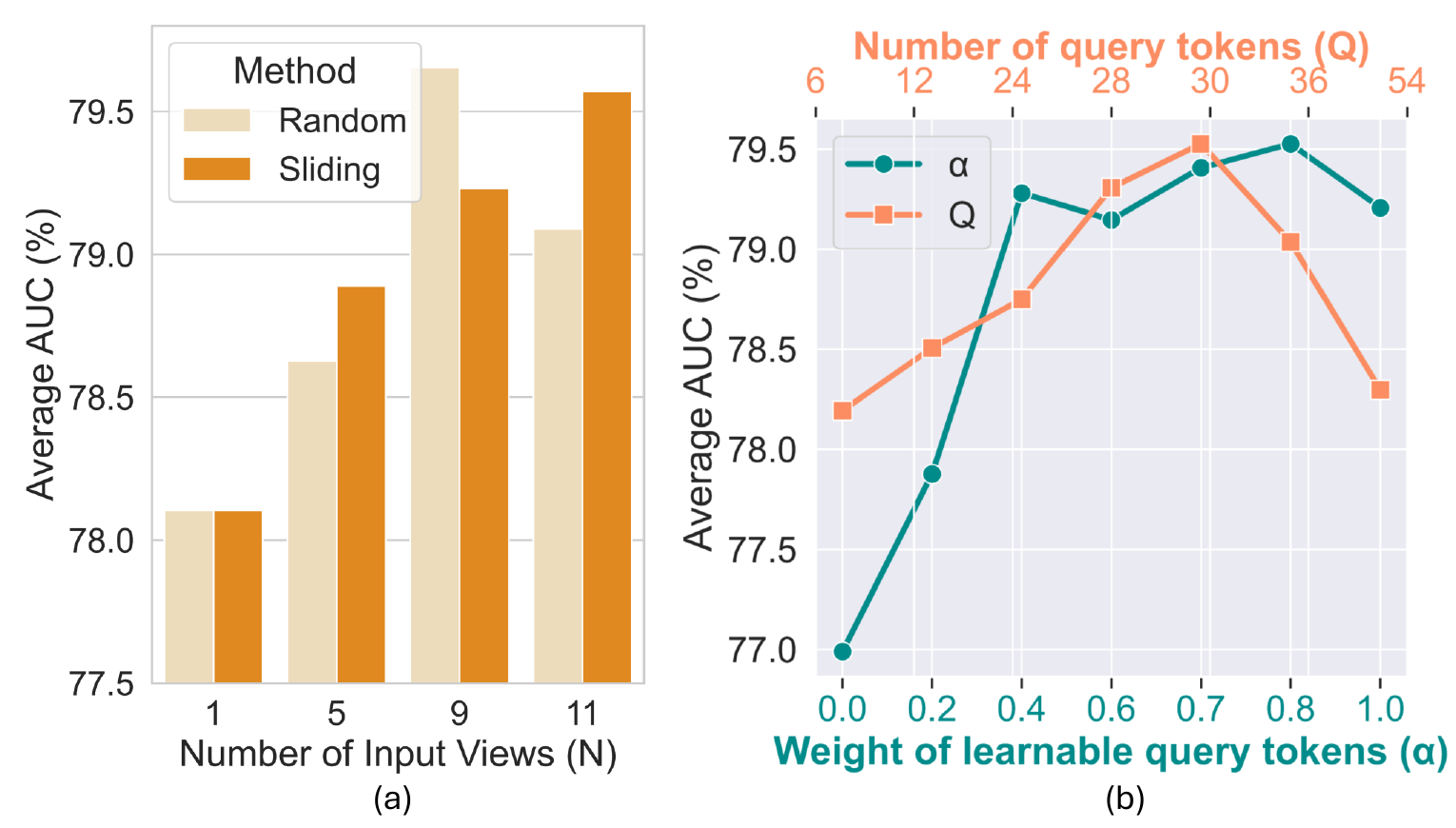}
\vspace{-1.5em}
    \caption{Ablation on (a) multi-view sampling and (b) query number $Q$ and weight factor $\alpha$. We report average AUC on the \emph{validation sets} of three representative datasets: CheXpert, MIMIC-CXR, and RSNA.}
    \vspace{-.8em}
    \label{fig:input_crop}
\end{wrapfigure}
\noindent \textbf{Multi-view sampling}: \cref{fig:input_crop}a explored varying the number of input views ($N$) in MQEncoder. Two strategies are reported: randomly sampling the local regions or systematically extracting them using a sliding window with overlap. 
Using only the global resized image (single-view, $N=1$) led to around 1.7\% drop in AUC from peak performance, while adding more views improved accuracy. \\
\noindent\textbf{Weighting factor \(\alpha\)}: 
This constant factor controls the contribution of learnable queries from MQEncoders for the diagnosis. \cref{fig:input_crop}b shows that \(\alpha\) within [0.4, 1.0] yields stable performance, while a near-zero value (a minimal contribution) leads to a substantial drop in AUC for 2.5\%. \\
\noindent\textbf{Query tokens}: 
\cref{fig:input_crop}b shows the effect of varying the number of learnable query tokens (\(Q\)) in MQEncoders (details in the appendix 
). Considering a 1\% margin of the performance peak, \(Q\) in the range of [24, 36] yields improved results, corresponding to around 10–18\% of the frozen tokens. 
The result in the appendix further shows that a shallow-to-moderate injection depth ($n_2 = 3$) provides the best trade-off between diagnostic accuracy and computational efficiency.\\
\noindent\textbf{Removing key components} to isolate their individual contribution.
\cref{tab:moe} (lower) shows that removing learnable queries causes the largest drop ($-8.09\%$ AUC on CheXpert, $-7.45\%$ on RSNA), confirming that query injection is the most critical for closing the domain gap. Removing the MoE and multi-view sampling causes consistent degradations of around $-5\%$ and $-3.5\%$. It validates that MedBridge's performance arises from its synergistic interaction rather than any single design choice.
\vspace{-0.5em}

\section{Discussion and Conclusion}
\label{sec:conclusion}
\vspace{-0.5em}

We introduced MedBridge, a lightweight domain adaptation framework designed to flexibly adapt foundation VLMs for multi-label diagnosis in chest X-rays.  
Unlike costly medical domain-specific pretraining that requires days of multi-GPU computation~\cite{tiu2022chexzero}, MedBridge completes adaptation in a few hours on a single GPU 
(Appendix)
while consistently improving diagnostic performance. 
Across five datasets in three key evaluations,
MedBridge achieves strong gains in accuracy, data efficiency, and cross-dataset generalization over existing adaptation methods.
These gains stem from its synergistic design, in which learnable queries act as a compact interface for domain alignment, fine-grained pathology preservation, and heterogeneous foundation VLMs routing jointly.
MedBridge is also model-agnostic and can readily adapt arbitrary foundation VLMs, with performance scaling effectively when integrating larger, more specialized models, such as LLaVA, Qwen-VL, and MedGemma.


\section*{Acknowledgements}

This work was supported by the Munich Center for Machine Learning (MCML), the German Research Foundation (DFG), and the DAAD programme Konrad Zuse Schools of Excellence in Artificial Intelligence, sponsored by the Federal Ministry of Research, Technology and Space. The authors gratefully acknowledge the scientific support and resources of the AI service infrastructure LRZ AI Systems provided by the Leibniz Supercomputing Centre (LRZ) of the Bavarian Academy of Sciences and Humanities (BAdW), funded by Bayerisches Staatsministerium für Wissenschaft und Kunst (StMWK).

\clearpage
\newpage
{
\small
\bibliography{main}
}








\end{document}